\documentclass[accepted]{article}

\usepackage{microtype}
\usepackage{graphicx}
\usepackage{subfigure}
\usepackage{tabularx}
\usepackage{subfiles}
\usepackage{multirow}

\usepackage{hyperref}

\usepackage{icml2024}
\usepackage[T1]{fontenc}

\usepackage{amsmath}
\usepackage{amssymb}
\usepackage{mathtools}
\usepackage{amsthm}
\usepackage[capitalize,noabbrev]{cleveref}

\icmltitlerunning{Transformer-Based Approaches for Sensor-Based Human Activity Recognition: Opportunities and Challenges}

\begin{document}

\twocolumn[
\icmltitle{Transformer-Based Approaches for Sensor-Based Human Activity Recognition: Opportunities and Challenges}



\icmlsetsymbol{equal}{*}

\begin{icmlauthorlist}
\icmlauthor{Clayton Souza Leite}{}
\icmlauthor{Henry Mauranen}{}
\icmlauthor{Aziza Zhanabatyrova}{}
\icmlauthor{Yu Xiao}{}\\
Aalto University, Finland
\end{icmlauthorlist}




\vskip 0.3in
]




\begin{abstract}

Transformers have excelled in natural language processing and computer vision, paving their way to sensor-based Human Activity Recognition (HAR). Previous studies show that transformers outperform their counterparts exclusively when they harness abundant data or employ compute-intensive optimization algorithms. However, neither of these scenarios is viable in sensor-based HAR due to the scarcity of data in this field and the frequent need to perform training and inference on resource-constrained devices. Our extensive investigation into various implementations of transformer-based versus non-transformer-based HAR using wearable sensors, encompassing more than 500 experiments, corroborates these concerns. We observe that transformer-based solutions pose higher computational demands, consistently yield inferior performance, and experience significant performance degradation when quantized to accommodate resource-constrained devices. Additionally, transformers demonstrate lower robustness to adversarial attacks, posing a potential threat to user trust in HAR.

\end{abstract}

\section{Introduction} \label{sec:intro}

Transformers \cite{transformers} have proven remarkably effective in natural language processing (NLP) \cite{gpt, bert}. Their success is mainly attributed to more effective learning of long-range dependencies in sequences without suffering from the vanishing gradient problem. Additionally, their capacity for training parallelism contributes to their prevalence. Recent advancements in computer vision (CV) \cite{vit, cvtransformer, cavit} also provide compelling evidence for the wide adoption of transformer architectures. Transformers have made their way to sensor-based Human Activity Recognition (HAR) \cite{vanillatransformerhar, ttn, hart, tasked}. Motivated by the more effective learning of long-range temporal dependencies, Mahmud et al. \cite{vanillatransformerhar} introduced a transformer-based network for HAR. The authors drew an analogy between data samples and words, as well as between windows and sentences to motivate transformers in HAR.

Transformers depart from the induction biases present in convolution and recurrent layers in favor of a general-purpose architecture. This poses a challenge. To outperform their counterparts, in NLP and CV, transformers require additional efforts \cite{flatminimavit} in the form of extensive pre-training \cite{vit, vivit, cvtransformer}, strong data augmentations \cite{strong_dataaug}, availability of excessive amounts of data \cite{transformers_medicalfield, vit_imagerestoration}, or more computationally-intensive optimization algorithms \cite{flatminimavit}. This results in larger data demands, increased computational efforts, and the need for excessive hyper-parameter fine-tuning \cite{flatminimavit}. In HAR, these endeavors have not been explored given the scarcity of data in this field \cite{moredatahar}. Instead, \citet{vanillatransformerhar}, \citet{ttn}, \citet{hart}, and \citet{tasked} introduced inductive biases into transformer-based models for HAR by incorporating convolutions. Nevertheless, the use of convolutions remains minimal, as their methods predominantly rely on the induction-free characteristics inherent in transformers. Hence, the rationale for applying transformers in HAR is questionable.

Transformers have been known to increase inference costs \cite{retnet, song2022xvit} in comparison to models relying on recurrent layers. This poses a challenge in the context of HAR, where model inference is often executed on edge devices with considerable resource constraints. Transformers have also demonstrated a tendency to converge to sharper minima \cite{flatminimavit}, which is associated with sub-optimal generalization and increased sensitivity to minor variations in the parameter space \cite{flatminima, flatminima2}. In HAR applications, these minor parameter variations are introduced through quantization procedures, which reduces the precision of the model's parameters and activations in favor of reduced inference costs in terms of both computational effort and memory footprint. 

The existence of large eigenvalues in the Hessian eigenspectrum, as observed in transformers \cite{flatminima}, has been found to correlate with poor robustness against adversarial examples \cite{adversarial_robustness, adversarial_robustness2}. Increased vulnerability to adversarial examples can jeopardize trust in fields like HAR for healthcare (e.g., misclassification of a fall in a fall detection system can have severe and potentially fatal consequences) and may result in increased performance variability under diverse and dynamic real-world scenarios. 

Based on the aforementioned concerns, we express skepticism regarding the applicability of transformers for HAR. We aim to take a closer look at the claimed superior performance of transformers for HAR in contrast to architectures relying solely on recurrent and convolutional layers. Employing techniques such as loss landscape visualization and Hessian analysis, we seek to obtain insights into the generalization capabilities of transformers in HAR. We also investigate the computational cost associated with transformer-based architectures during inference and training, examining the impact of quantization (both with and without post-quantization training). Additionally, we explore the susceptibility of transformers to adversarial examples in the context of HAR.

To address the challenges of convergence to sharp minima and the prevalence of large-magnitude eigenvalues in transformers, the Sharpness-Aware Minimization (SAM) \cite{sam, flatminimavit} method has been proposed as an optimization strategy. To reduce the inference costs associated with transformer networks, \cite{retnet} introduced Retentive Networks (RetNet). This architecture combines the training parallelism advantages from transformers with lower computational costs during inference, similar to recurrent layers. RetNet is capable of achieving performance similar to transformers. We also evaluate these countermeasure solutions in transformers for HAR. Our findings are summarized as follows.

\begin{itemize}
    \item Transformer-based models show consistently lower performance than non-transformers. Transformers can only outperform their counterparts in less than 3\% of the experiments and show higher difficulty learning more complex activities.
    \item Transformers in HAR suffer from sharp minima, and a moderate correlation exists between sharpness and performance generalization. SAM improves the performance of transformers. Yet, not sufficiently as they are still outperformed by their counterparts. Furthermore, our observations reveal a correlation between sharpness and the degradation of performance after quantization. Architectures that converge to sharper minima experience extreme performance degradation after quantization. These performance losses can reach up to 85\%, whereas non-transformer models typically remain under 5\% performance degradation.
    \item Transformers are more susceptible to adversarial attacks, in special architectures that completely abdicate from the inductive bias present in convolutions or recurrent layers.
    \item Training transformers demands significantly more computational resources, ranging from 2.5 to 26 times more effort. While some transformer models may exhibit faster inference times, such as RetNet, their performance is significantly inferior.
\end{itemize}


The rest of the paper is structured as follows. Section 2 presents the background and related work, followed by the methodology in Section 3. Section 4 contains the conducted experiments with their respective analyses. Section 5 concludes the work.

\section{Background and Related Work} \label{related}


\begin{figure*}
    \centering
    \includegraphics[width=0.85\textwidth]{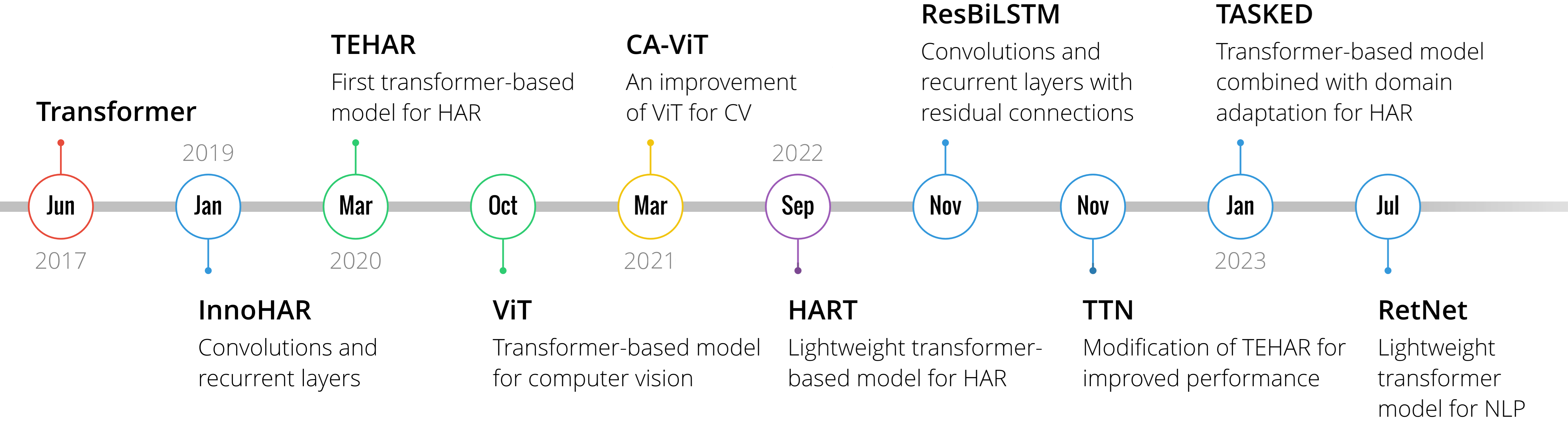}
    \caption{Timeline of the studied neural network architectures. InnoHAR, TEHAR, HART, ResBiLSTM, TTN, and TASKED are in the field of HAR. With the exception of TTN which improves over TEHAR, these architectures have been developed in a scattered manner, lacking a clear interrelation and comparison among each other. With the exception of the original Transformer, we implement all the nine depicted neural network architectures.}
    \label{fig:timeline}
\end{figure*}


 
In recent years, substantial research efforts have focused on advancing deep learning-based HAR algorithms, primarily by proposing ever more sophisticated neural network architectures for higher performance. Earlier efforts involved simply stacking convolutional and recurrent layers \cite{ordonez} or solely relying on LSTM (Long Short-Term Memory) layers \cite{ensembleslstm}. More recently, InnoHAR \cite{innohar} combined the concept of inception neural networks with recurrent neural networks. Specifically, \textbf{InnoHAR} is composed of four stacked inception-like modules followed by two Gated Recurrent Unit (GRU) layers before a dense layer with a softmax activation function for classification.  \textbf{ResBiLSTM} \cite{resbilstm} utilized a residual block with convolutions to extract spatial features from the input data and a bi-directional LSTM for temporal feature extraction. It also utilizes a residual connection at the level of convolutions.

Given the notable success of transformers in NLP and CV, \citet{vanillatransformerhar} introduced a transformer neural network for HAR (\textbf{TEHAR}).  It first extracts attention scores for each individual sensor channel, providing a weighted version of the input (sliding window) determined by the learned contributions of the sensor channels. Subsequently, positional encoding is applied to the weighted input and is processed in a standard Transformer encoder, resulting in a 3D tensor (with the batch dimension included). Prior to the classification layer, the tensor undergoes a weighted average operation along the temporal dimension. The significance of each element in the temporal dimension is determined through dense layers, allowing for the adaptation to the learned temporal dynamics. Compared to its counterparts (e.g., DeepConvLSTM by \citet{ordonez}) in four distinct datasets (i.e., Opportunity, Skoda, PAMAP2, and USC-HAD), TEHAR demonstrated superior performance. However, the code released by \citet{vanillatransformerhar} contained a significant error in which the labels were included as input data. The error has been solved in our implementation of TEHAR. 

Two-Stream Transformer Network (\textbf{TTN}) \cite{ttn} was proposed as an improvement to TEHAR. The method involved processing input data in two distinct streams to extract spatial and temporal features separately. The neural network’s input undergoes parallel processing through both streams. Before the transformer layers, in the spatial stream, sensor attention is applied, whereas, in the temporal stream, positional encoding is utilized. The outputs of these streams are concatenated prior to the classification layer. \textbf{HART} \cite{hart} introduced lightweight transformer-based architectures. HART divides the input data along the channel (sensor) dimension, subjecting each split to convolutions that reduce the dimensionality prior to transformer layers.
Distinct transformer-based layers are employed for each split, generating individualized features which are then concatenated before the classification layer. However, their evaluation did not include a comparison with state-of-the-art non-transformer-based architectures. 



Transformer-based Adversarial learning for human activity recognition using wearable sensors via Self-KnowledgE Distillation (TASKED) \cite{tasked} addresses the prevailing generalization challenges encountered in sensor-based HAR methods. \textbf{TASKED} encoder is built from a transformer and a bypassing single-layer convolutional channel. The architecture implements a subject discriminator, Maximum Mean Discrepancy (MMD) loss based on subjects, and adversarial training with knowledge distillation between a pre-trained and a primary model. This trains a subject-agnostic latent space for classification. However, in TASKED, the absence of experiments that integrate their domain adaptation approach with non-transformer-based architectures leaves a gap in explaining the specific advantages and relevance of transformers in the context of HAR. Additionally, the higher reported performance by \citet{tasked} required extra training efforts, involving the use of test data in an unsupervised manner. In HAR, data are severely scarce when compared to fields such as NLP and CV \cite{moredatahar}. Figure \ref{fig:timeline} depicts the timeline of the aforementioned architectures. 



\section{Methodology} \label{methodology}

In this section, we introduce the neural network architecture selection and the implementation. We also describe the tools employed for analyzing neural network architectures: loss landscape, Hessian eigenvalues, quantization, adversarial attacks, and SAM. The first two tools aim to offer insights into the trainability characteristics of the architectures. Quantization and adversarial attacks serve to gauge the robustness of the models against slight variations in parameters and inputs, respectively. Lastly, SAM is an optimization algorithm known to enhance trainability and improve robustness against fluctuations in parameters and inputs. We will also evaluate its impact on transformers for HAR. Subsequently, we describe the datasets utilized in our analyses and how the entire analysis framework was implemented in code.

\subsection{Model Selection and Implementation}

We implement several neural network architectures, including ResBiLSTM \cite{resbilstm}, InnoHAR \cite{innohar}, TEHAR \cite{vanillatransformerhar}, TTN \cite{ttn}, HART \cite{hart}, and TASKED \cite{tasked}. Our implementation of TASKED omits the adversarial training and subject discriminator to focus on the transformer structure and keep the training methodology in line with other implemented models. Additionally, we adapt Vision Transformer (ViT) \cite{vit}, Cross-Attention Vision Transformer (CA-ViT) \cite{cavit}, and RetNet \cite{retnet} to the domain of HAR.

\textbf{ViT} was originally employed for image input data. It splits the image into fixed-size patches, which are flattened and linearly projected before receiving position encoding. The embeddings are then passed through the standard Transformer encoder before the classification layer. ViT can be employed for HAR by regarding sliding windows as 3D tensors with shape \textit{(window length, number of sensor channels, 1)}. \textbf{CA-ViT} extends the ViT concept by incorporating two parallel streams: the primary and the complementary streams. Both streams operate by splitting the image into fixed-size patches. The primary stream utilizes larger-sized patches compared to the complementary one. Each stream then independently undergoes the processes of flattening and linear projection, followed by the standard Transformer encoder. Finally, \textbf{RetNet} has a dual mode. During training, the architecture closely resembles the standard transformer, facilitating training parallelism. During inference, the architecture transforms into a recurrent model, providing reduced inference costs.

In ViT and CA-ViT, we treat a sliding window as a grayscale image, with the height (width) dimension corresponding to the time (sensor) dimension within the window. For RetNet, we incorporate a Conv1D layer before the retention layers, serving as a tokenization method. The training of transformer-based architectures (i.e., TEHAR, TTN, TASKED, ViT, CA-ViT, and RetNet) involves both the Adam optimizer and the combination of SAM with Adam. Following the completion of training, we conduct a comprehensive analysis of the models based on their recognition performance, topology of the loss landscape, Hessian eigenvalues, robustness to adversarial attacks, performance degradation due to quantization, and training and inference costs.


To measure the performance of the neural network, we utilize the weighted F1-score ($F_w$). The F1-score for a certain class is defined as $ \frac{2 \text{TP}}{2 \text{TP} + \text{FP} + \text{FN}}$, in terms of true positives (TP), false positives (FP), false negatives (FN), and true negatives (TN). $F_w$ employs a weighted average across all classes, where the weight assigned to each class is defined by the proportion (or frequency) of that class in the dataset.

\subsection{Loss Landscape}
\label{sec:losslandscape}

The loss landscape (or surface) refers to the topography of the loss function with respect to the model's parameters $\pmb \theta$. It provides insights into the model's ability to generalize to unseen data. During the training process, the objective is to converge towards flatter minima since research suggests that such minima often signify better generalization capability, as opposed to sharper minima \cite{flatminima, flatminima2}. The correlation between the sharpness of minima and specific neural network architectures has been demonstrated in prior studies \cite{flatminimavit}. Hence, by visualizing the loss landscape, our goal is to extract insights regarding the generalization performance of transformers for HAR.

To visualize the loss landscape, a fixed point in the parameter space $\pmb \theta'$ and two direction vectors $\pmb \delta$ and $\pmb \eta$ are considered. One then plots the function of the form given in Eq. \ref{eq:losslandscape} \cite{losslandscape}.

\begin{equation}
f(\alpha, \beta) = L(\pmb X, \pmb Y, \pmb \theta' + \alpha \cdot \pmb \delta + \beta \cdot \pmb \eta),
\label{eq:losslandscape}
\end{equation}

where $L(\cdot)$ is the loss function (e.g., cross-entropy in the classification case) calculated for a batch of examples $\pmb X$ and their respective labels $\pmb Y$. $\alpha$ and $\beta$ are scalars that scale the direction vectors $\pmb \delta$ and $\pmb \eta$, respectively, to perturb the model's parameters. We apply the approach developed by \citet{losslandscape} to find the direction vectors $\pmb \delta$ and $\pmb \eta$. In their approach, each direction vector has the same dimensions are the model's parameters (i.e., $\pmb \theta$) and is drawn from the Gaussian distribution with unit mean and identity covariance matrix. Since  \citet{losslandscape} devised the method considering neural networks with only convolutional layers, the direction vectors can be interpreted as convolutional filters. Each convolutional filter in the direction vectors is normalized according to Eq. \ref{eq:dir_vector}.

\begin{equation}
     \delta_{i, j} \leftarrow \frac{  \delta_{i, j} }{ ||  \delta_{i, j} ||} \cdot || \theta_{i, j} ||,
      \label{eq:dir_vector}
\end{equation}

where $\delta_{i, j}$ refers to the $j$\textit{-th} convolutional filter of layer $i$\textit{-th}. Similarly, $\theta_{i, j}$ represents the weights of the  $j$\textit{-th} convolutional filter of layer $i$. The function $||\cdot ||$ is the Frobenius norm. The concept can be easily expanded for dense layers since they can be interpreted as convolutional layers with 1 x 1 output feature maps and the number of filters equal to the number of neurons. Similarly, the weight matrices of recurrent layers can be interpreted as dense layers. To generate the loss landscapes, we vary the parameters $\alpha$ and $\beta$ (refer to Eq. \ref{eq:losslandscape}) across a range from -3 to 3 with a step size of 0.2.

\subsection{Hessian Analysis}

Hessian analysis \cite{hessianeigenvaluedensity} shares a lot of similarities with the analysis of the loss landscape. Hessian is the matrix of second-order partial derivatives of a function. The Hessian matrix is extremely large in deep learning contexts, and an efficient way to estimate the density of Hessian eigenvalues was developed by \citet{hessianeigenvaluedensity}, which we apply in this study. We define the order parameter to 50. This parameter dictates the precision of the approximation of Hessian eigenvalues. The value of 50 was found to deliver good precision with reduced resource costs.

As Hessian describes the rate of second-order increase along different parameters, the eigenvectors and eigenvalues of the matrix give the directions and rates of increase with the greatest magnitudes. In deep learning, we end up with a mixture of positive and negative eigenvalues, i.e., saddle points. The aim of the analysis is then to interpret the distribution of these eigenvalues to get a sense of the local sharpness of the loss surface, instead of adhering strictly to the interpretation of classical optimization. 

\subsection{Quantization}

Quantization refers to the reduction in precision (e.g., from 32-bit floating point to 16-bit floating point or 8-bit integer representation) of the model's parameters and activations for lower memory utilization and higher computational efficiency. Quantization can lead to performance degradation, which can be mitigated through a post-training procedure that leverages a small set of data samples. To analyze the impact of the quantization, we convert the initially trained models from 32-bit floating point to 8-bit integer representations using TensorFlow Lite. Post-quantization training is also applied by utilizing a subset of the training set containing 256 input-output pairs.

\subsection{Adversarial Attacks}

Adversarial attacks are a form of manipulating the input space of a neural network in order to change the outputs \cite{adversarialattacks}. Here we focus on one of the classical forms of adversarial attacks, the Fast Gradient Sign Method (FGSM) \cite{adversarialattacks}, which is defined as $x_{adv} = x + \epsilon \cdot \textrm{sign}(\nabla_x L(x, y, \theta))$. The goal is to find adversarial examples $x_{adv}$ by adding a small ($\epsilon$) sized perturbation along the gradient of the loss function $L(x, y, \theta)$ with respect to the input $x$ and its class $y$. The same step is suggested to be used as a regularization term by \citet{adversarialattacks}. The main purpose of this method is to analyze the sharpness of the loss surface in input space and correct it as either a regularization term or by including the adversarial examples as part of the training set. We perform experiments with the parameter $\epsilon$ set to $0.01$.

\subsection{Sharpness-Aware Minimization}

As stated in Section \ref{sec:losslandscape}, one desires to reach flatter minima for enhanced generalization. In line with this, \citet{sam} proposed Sharpness-Aware Minimization  (SAM) -- an optimization technique that seeks to minimize not only the loss itself but also the sharpness of the loss, thereby facilitating convergence towards flatter minima. Rather than computing the gradient at the current point $\pmb \theta'$ in the model's parameter space, SAM introduces a perturbation term $\pmb \epsilon(\pmb X, \pmb Y, \pmb \theta')$ to $\pmb \theta'$. This is defined in Eq. \ref{eq:sam} and Eq. \ref{eq:sam2}.

\begin{equation}
    \nabla_{\pmb \theta} L^{SAM}(\pmb X, \pmb Y, \pmb \theta) = \nabla_{\pmb \theta} L(\pmb X, \pmb Y, \pmb \theta) |_{ \pmb \theta = \pmb \theta' + \pmb \epsilon(\pmb X, \pmb Y, \pmb \theta')}
\label{eq:sam}
\end{equation}

\begin{equation}
    \pmb \epsilon(\pmb X, \pmb Y, \pmb \theta') = \rho \text{ } \nabla_{\pmb \theta} L(\pmb X, \pmb Y, \pmb \theta) ) \big/ \|\nabla_{\pmb \theta}  L(\pmb X, \pmb Y, \pmb \theta) \|,
    \label{eq:sam2}
\end{equation}

where $\rho$ is a positive constant. We employ SAM in the training of transformer-based architectures and assess their effects on performance in comparison to scenarios where it is not utilized. 

\subsection{Datasets}

\noindent
\textbf{Opportunity (O).} The Opportunity dataset \cite{opportunity} consists of 18 kitchen-related activities recorded with body-worn, object, and ambient sensors. Four users participated in the data collection and their activities were annotated at different levels. Namely, four modes of locomotion (i.e., standing, walking, sitting, and lying down), 17 mid-level activities (e.g., opening a door), and five high-level activities (e.g., preparing breakfast). These three variants are denoted as OL, OM, and OH, respectively.


\noindent
\textbf{PAMAP2 (P).} The PAMAP2 dataset \cite{pamap2} is formed by daily-life activity data collected from nine participants who wore a heart rate monitor and three Inertial Measurement Units (IMUs). The dataset initially included 18 daily life activities. However, in alignment with prior studies \cite{ensembleslstm, ttn}, six rare activities were omitted during training to prevent heavy imbalance. The null class, which represents transient activities, is also excluded. 


\noindent
\textbf{SKODA (S).} The dataset \cite{skoda} encompasses 10 activities associated with a car assembly line scenario. The data were collected from a single participant equipped with five accelerometers on each arm. 


\noindent
\textbf{USC-HAD (UH).}  This dataset \cite{uschad} encompasses 12 daily-life activities acquired from 14 participants equipped with an accelerometer and a gyroscope at the waist. 


\begin{table}[tb]
\small
\centering
\begin{tabular}{lcccccc}
\hline
 &
  \multicolumn{1}{l}{\textbf{OH}} &
  \multicolumn{1}{l}{\textbf{OM}} &
  \multicolumn{1}{l}{\textbf{OL}} &
  \multicolumn{1}{l}{\textbf{P}} &
  \multicolumn{1}{l}{\textbf{S}} &
  \multicolumn{1}{l}{\textbf{UH}} \\ \hline \hline
Timesteps per window    & 256 & 32  & 32  & 32 & 32 &  32 \\ \hline
Overlap timesteps       & 128 & 16  & 16  & 16 & 16 &  16\\ \hline
Number of classes       & 5   & 18  & 4   & 12 & 10 & 12\\ \hline
Number of channels      & 240 & 133 & 133 & 52 & 60 & 6  \\ \hline
Number of subjects      & 4 & 4 & 4 & 9 & 1 & 14\\ \hline
\end{tabular}
\caption{Additional details regarding the datasets. PAMAP2, USC-HAD, and Skoda were decimated from their original sampling frequency to 33Hz. Opportunity's original sampling frequency is 33Hz.}
\label{tab:details_datasets}
\end{table}

Table \ref{tab:details_datasets} includes additional information on the datasets. For the Opportunity dataset, the validation set comprised run 2 of subject 1, while the test set encompassed runs 3 and 4 of subjects 2 and 3. In the case of PAMAP2, subjects 5 and 6 were designated as the validation and test sets, respectively. The validation and test sets for SKODA each included 15\% of the total data selected randomly. Regarding USC-HAD, subjects 11 and 12 constituted the validation set, while subjects 13 and 14 formed the test set. The choice of validation and test sets are in accordance with previous studies \cite{ttn, vanillatransformerhar, innohar, resbilstm, tasked}.

\subsection{Code Implementation and Execution}

Our analysis framework, together with the distinct architectures, was implemented in Python 3.8.10 utilizing the TensorFlow 2.12.0 version. The code was executed on an NVIDIA Tesla V100 with 16GB of video memory and an Intel(R) Xeon(R) Gold 6134 CPU @ 3.20GHz. The training process was controlled by setting a maximum of 20 epochs. The hyper-parameter selection process was conducted through a trial-and-error approach, with the performance on the validation set as the evaluation metric. Prior to the training, the data were normalized to zero mean and unit variance.


\section{Experimental Results} \label{experiments}

\begin{table}[tb]
\small
\centering
\begin{tabular}{lcccccc}
\hline
 &
  \multicolumn{1}{c}{\textbf{OH}} &
  \multicolumn{1}{c}{\textbf{OM}} &
  \multicolumn{1}{c}{\textbf{OL}} &
  \multicolumn{1}{c}{\textbf{P}} &
  \multicolumn{1}{c}{\textbf{S}} &
  \multicolumn{1}{c}{\textbf{UH}} \\ \hline \hline
ResBiLSTM       & \textbf{0.820} & \textbf{0.892} & \textbf{0.906} & 0.882 & 0.964 & \textbf{0.629} \\ \hline
InnoHAR         & 0.796 & 0.891 & 0.901 & 0.846 & 0.951 & 0.596 \\ \hline 
ViT             & 0.697 & 0.878 & 0.890 & 0.842 & 0.964 & 0.481  \\ \hline
ViT-SAM         & 0.766 & 0.890 & 0.898 & 0.881 & \textbf{0.983} & 0.610  \\ \hline
CA-ViT          & 0.677 & 0.847 & 0.823 & 0.825 & 0.932 & 0.439  \\ \hline
CA-ViT-SAM      & 0.716 & 0.850 & 0.858 & 0.824 & 0.875 & 0.400  \\ \hline
TEHAR           & 0.710 & 0.838 & 0.877 & 0.817 & 0.837 & 0.506 \\ \hline
TEHAR-SAM       & 0.787 & 0.867 & 0.880 & 0.836 & 0.825 & 0.557 \\ \hline
TTN             & 0.771 & 0.859 & 0.894 & 0.829 & 0.920 & 0.510 \\ \hline
TTN-SAM        & 0.789 & 0.848 & 0.891 & 0.815 & 0.912 & 0.615 \\ \hline
RetNet         & 0.776 & 0.846 & 0.859 & 0.797 & 0.884 & 0.507  \\ \hline
RetNet-SAM  & 0.730 & 0.862 & 0.856 & 0.778 & 0.903 & 0.427  \\ \hline
TASKED  & 0.709 & 0.870 & 0.879  & 0.802 & 0.903  &  0.518 \\ \hline
TASKED-SAM  & 0.767 & 0.867 &  0.891 & \textbf{0.925} & 0.891  &  0.474 \\ \hline
HART  & 0.692  & 0.866  & 0.860   & 0.717  & 0.942   &  0.622  \\ \hline
HART-SAM  & 0.739  & 0.871  & 0.877   & 0.773  & 0.896   &  0.602  \\ \hline
\end{tabular}
\caption{Experimental results of the neural network architectures in terms of weighted F1-score $F_w$. }
\label{tab:results1}
\end{table}

\begin{figure*}
    \centering
    \includegraphics[width=1\linewidth]{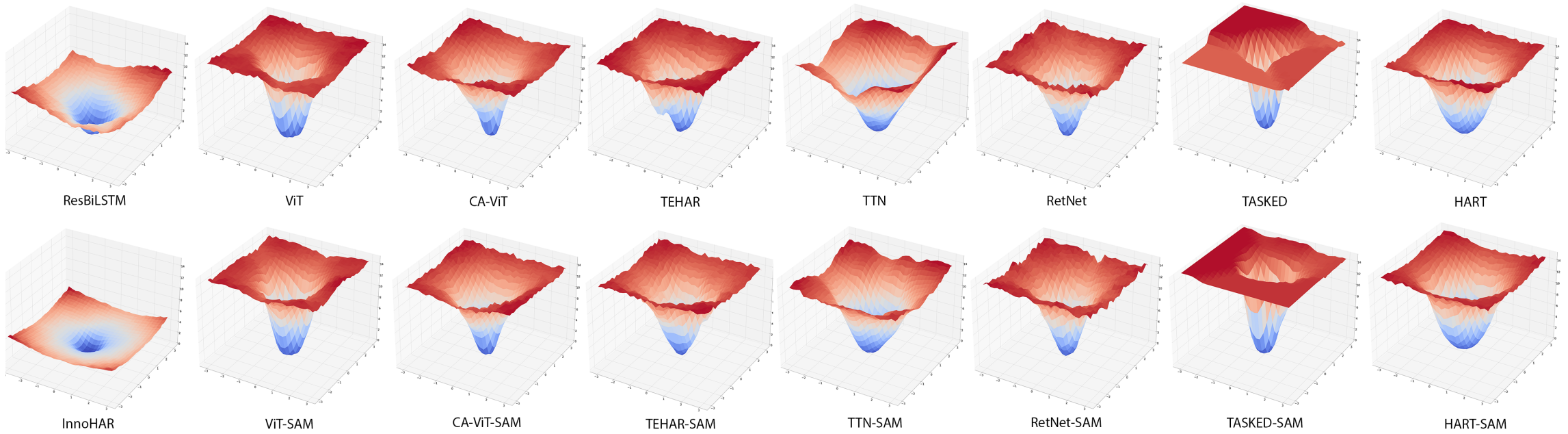}
    \caption{Loss landscapes of the trained architectures for PAMAP2. The x and y axes vary from -3 to 3. The z-axis ranges from 0 to 15.}
    \label{fig:ll}
\end{figure*}

\begin{figure*}
    \centering
    \includegraphics[width=0.95\linewidth]{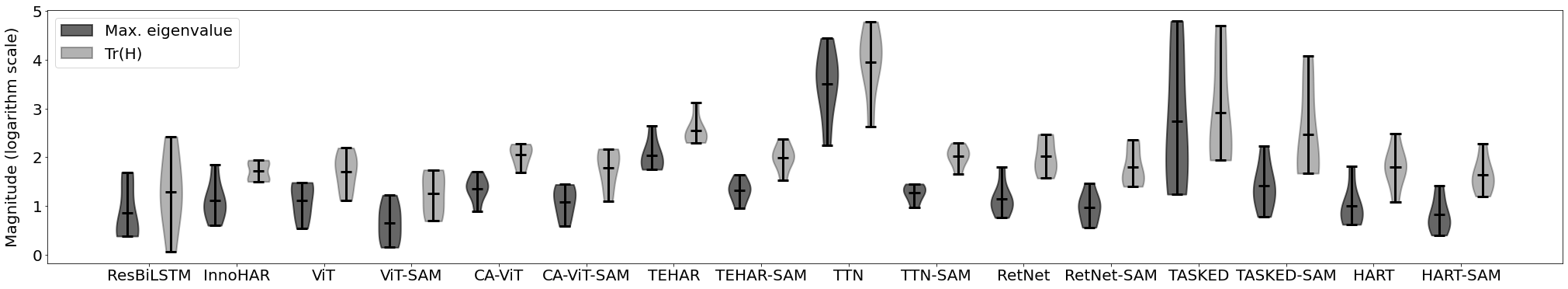}
    \caption{Trace of the Hessian and maximum eigenvalue across all datasets. The y-axis is expressed in a logarithm scale. Lower is better.}
    \label{fig:hessians}
\end{figure*}

\textbf{Transformer-based models show consistently lower performance than ResBiLSTM and InnoHAR.} Table \ref{tab:results1} includes the performance results of the tested architectures across all datasets. When averaging across all datasets, the top five architectures in descending order of performance are ResBiLSTM, InnoHAR, TTN, ViT, and HART. ResBiLSTM outperforms the leading transformer model (TTN) by a margin exceeding 5\%. Among all the 84 experiments (six datasets times 14 architectures) conducted with transformer-based models, ResBiLSTM is outperformed only twice (by ViT-SAM in SKODA and TASKED-SAM in PAMAP2). InnoHAR, on the other hand, is outperformed in six out of the 84 experiments. 

\textbf{Transformer-based models exhibit larger difficulty in learning tasks of higher complexities.} For each dataset, the further the performance ($F_w$ score) of ResBiLSTM from 1, the larger its disparity to the average performance of transformer models. This contrast is exemplified by datasets like OpportunityLoco, OpportunityML, and OpportunityHL, which exhibit increasing levels of complexity despite having the same data distribution. This contradicts the fundamental motivation for employing transformers in HAR introduced by \citet{vanillatransformerhar}. Namely, the belief that transformers excel at learning spatial and temporal dependencies from sensor data, especially long-range dependencies present in complex activities.

\textbf{Sharp minima are not caused by saturated neurons or activation functions.} Some earlier literature suggests that, in neural networks in general, saturated neurons and activation functions can contribute to sharp minima \cite{bosman_empirical_2023}. However, we observe this to not be the case in these transformer-specific sharp local minima. For a trained transformer-based architecture, the weights are appropriately distributed in each layer. Similarly, using different activation functions in the architecture did not change the sharpness of the found minimum after training. 

\textbf{SAM delivers performance improvement but does not suffice.} Except for RetNet, all transformer-based architectures demonstrate a performance boost averaging from 1\% (HART) to 4.6\% (ViT) across various datasets when trained with SAM. In the latter case, this improvement is significant, allowing ViT to outperform InnoHAR and approach ResBiLSTM with a 1\% margin on average. However, most transformer-based architectures still show significant performance gaps with respect to InnoHAR and ResBiLSTM.


\textbf{The effectiveness of SAM is influenced by the dataset.} SAM exhibits higher performance improvements on larger datasets such as Opportunity and PAMAP2. For the Skoda dataset, which is approximately 3-6 times smaller than Opportunity and PAMAP2, an average performance degradation of 1.3\% is observed when SAM is applied. This can be explained by the limited diversity per class of small datasets. SAM relies on accurate estimates of gradient sharpness to minimize loss sharpness. In datasets lacking diversity, these estimated sharpness values can be noisy, leading to ineffective weight updates. Regarding the Opportunity variants, we observed that SAM provides the most significant benefits to the least class-imbalanced dataset, OpportunityHL. In contrast, OpportunityML, where the null class dominates with approximately 70\%, shows a very small average improvement of less than 1\% with the application of SAM. This finding is in agreement with investigations by \citet{ccsam} and \citet{ccsam2}. In conclusion, SAM may not be a suitable choice in HAR due to significant data scarcity and frequent class imbalances, which jeopardizes the use of transformer-based architectures in this field.

\textbf{The reduced performance of transformers is linked to sharper minima.} Fig. \ref{fig:ll} illustrates the loss landscapes of the converged models in the PAMAP2 dataset. We omit the illustrations of the loss landscapes for the other datasets as no clear visual distinctions are evident across datasets. Fig. \ref{fig:hessians} illustrates the average trace of the Hessian and average maximum eigenvalue for each architecture. The average is calculated across the datasets. Transformer-based models exhibit significantly sharper minima, as evident in Fig. \ref{fig:ll} based on the landscape shape and in Fig. \ref{fig:hessians} through higher sharpness metrics. In select cases (TTN, TEHAR, TASKED), the measured sharpness is orders of magnitude higher compared to ResBiLSTM and InnoHAR.

\begin{table}[tb]
\centering
\small
\begin{tabular}{lcccccc}
\hline
 & \textbf{OH} & \textbf{OM} & \textbf{OL} & \textbf{P} & \textbf{S} & \textbf{UH} \\ \hline \hline
ResBiLSTM  & 3.95  & 2.33 & 3.72 & 1.94 & 2.71 & 3.36  \\ \hline
InnoHAR    & 7.04  & 3.20 & 2.58 & 1.83 & 4.24 & 2.78  \\ \hline
ViT        & 39.14 & 6.36 & 5.87 & 4.37 & 4.56 & 3.96  \\ \hline
ViT-SAM    & 36.58 & 4.55 & 3.80 & 3.62 & 4.97 & 3.62  \\ \hline
CA-ViT     & 30.17 & 3.12 & 4.78 & 2.64 & 7.90 & 2.55  \\ \hline
CA-ViT-SAM & 17.52 & 2.43 & 3.80 & 2.44 & 3.66 & 1.69  \\ \hline
TEHAR      & 3.90  & 1.89 & 1.36 & 2.23 & 5.08 & 4.06  \\ \hline
TEHAR-SAM  & 2.38  & 1.41 & 1.13 & 1.37 & 9.58 & 5.40  \\ \hline
TTN        & 6.98  & 3.51 & 3.21 & 1.18 & 3.67 & 4.09  \\ \hline
TTN-SAM    & 2.38  & 1.51 & 1.74 & 1.25 & 4.22 & 3.53  \\ \hline
RetNet     & 8.78  & 2.89 & 2.72 & 3.25 & 2.39 & 2.10  \\ \hline
RetNet-SAM & 12.69 & 1.93 & 2.33 & 2.95 & 3.70 & 1.55  \\ \hline
TASKED     & 6.77  & 2.11 & 2.51 & 0.11 & 5.70 & <0.01 \\ \hline
TASKED-SAM & 3.73  & 2.10 & 1.52 & 1.94 & 2.84 & <0.01 \\ \hline
HART       & 10.27 & 3.97 & 3.32 & 3.04 & 5.86 & 3.45  \\ \hline
HART-SAM   & 9.94  & 2.93 & 2.55 & 3.13 & 3.39 & 2.90  \\ \hline
\end{tabular}
\caption{Performance degradation in percentage points as a result of adversarial attacks. The magnitude of the perturbance $\epsilon$ was set to 0.01. Lower is better.}
\label{tab:resultsaa}
\end{table}

\begin{table}[tb]
\small
\centering
\begin{tabular}{lcccccc}
\hline
           & \textbf{OH} & \textbf{OM} & \textbf{OL} & \textbf{P} & \textbf{S} & \textbf{UH} \\ \hline \hline
ResBiLSTM  & 6.45        & 3.29        & 0.17        & 0.10       & 0.36       & 0.77        \\ \hline
InnoHAR    & 17.97       & <0.01       & 0.07        & 0.15       & 0.39       & 0.13        \\ \hline
ViT        & 34.12       & 0.23        & 0.22        & 0.47       & 18.94      & 2.26        \\ \hline
ViT-SAM    & 25.49       & <0.01       & 0.01        & 0.30       & 18.54      & <0.01       \\ \hline
CA-ViT     & 23.99       & 0.23        & <0.01       & 0.28       & 7.30       & <0.01       \\ \hline
CA-ViT-SAM & 22.32       & <0.01       & 0.38        & <0.01      & 4.63       & 0.25        \\ \hline
TEHAR      & 62.25       & 13.57       & 49.67       & 74.01      & 76.81      & 44.64       \\ \hline
TEHAR-SAM  & 62.52       & 12.21       & 51.75       & 74.43      & 71.95      & 49.56       \\ \hline
TTN        & 57.28       & 85.59       & 80.48       & 68.90      & 80.87      & 48.67       \\ \hline
TTN-SAM    & 56.06       & 53.17       & 51.54       & 73.92      & 82.95      & 48.20       \\ \hline
RetNet     & <0.01       & <0.01       & <0.01       & 0.60       & <0.01      & 8.38        \\ \hline
RetNet-SAM & 30.97       & <0.01       & <0.01       & <0.01      & <0.01      & 3.44        \\ \hline
HART       & 23.10       & 0.11        & <0.01       & 0.22       & 0.27       & 0.60        \\ \hline
HART-SAM   & 19.85       & <0.01       & 17.33       & <0.01      & 0.70       & 1.68        \\ \hline
\end{tabular}
\caption{Performance degradation in percentage points as a result of INT8 quantization. Lower is better.}
\label{tab:results2}
\end{table}

We find a moderate correlation between sharpness and generalization performance. Architectures such as TTN, ViT, TEHAR, and TASKED, which exhibit the most substantial performance improvements with SAM, also demonstrate the most significant reduction in sharpness, as depicted in Fig. \ref{fig:hessians}. Note that the correlation is not consistently valid across all cases. For instance, TASKED displays sharper minima compared to RetNet, CA-ViT, and TEHAR, but achieves better performance on average. This lack of strong correlation is in agreement with findings by \citet{similarhessian} and \citet{similarhessian2} and is not yet fully understood in the research community. 

\textbf{Transformers are more prone to adversarial attacks and suffer from higher post-quantization performance degradation.} Table \ref{tab:resultsaa} and Table \ref{tab:results2} present the performance degradation due to adversarial attacks and quantization, respectively. Note that TASKED is absent from the quantization experiments due to its incompatibility to be quantized by the TensorFlow Lite framework. Architectures that rely solely on transformer layers, such as ViT and CA-ViT, exhibit a lower robustness to adversarial attacks. This suggests that transformer layers may be the significant contributing factor. In contrast, TASKED, incorporating several convolutions, and TEHAR, which employs a combination of convolutions and fewer transformer layers, demonstrate higher robustness to adversarial attacks. SAM demonstrates limited effectiveness in mitigating adversarial attacks, with improvements typically under 1.5\%. 

\begin{figure}[tb]
    \centering
    \includegraphics[width=0.45\textwidth]{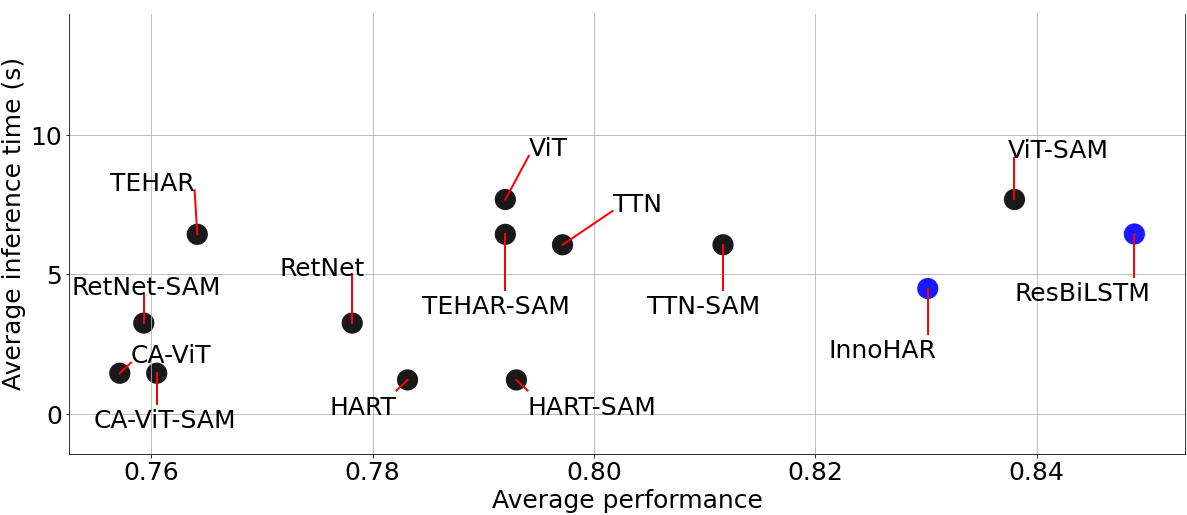}
    \caption{Average performance against inference time (in seconds) across all datasets, where inference time measures the duration required to infer all examples from the test set.  TASKED and TASKED-SAM achieve average performances of 0.780 and 0.802, respectively. Their inference time is 35.76s.}
    \label{fig:inference}
\end{figure}

\begin{figure}[tb]
    \centering
    \includegraphics[width=0.45\textwidth]{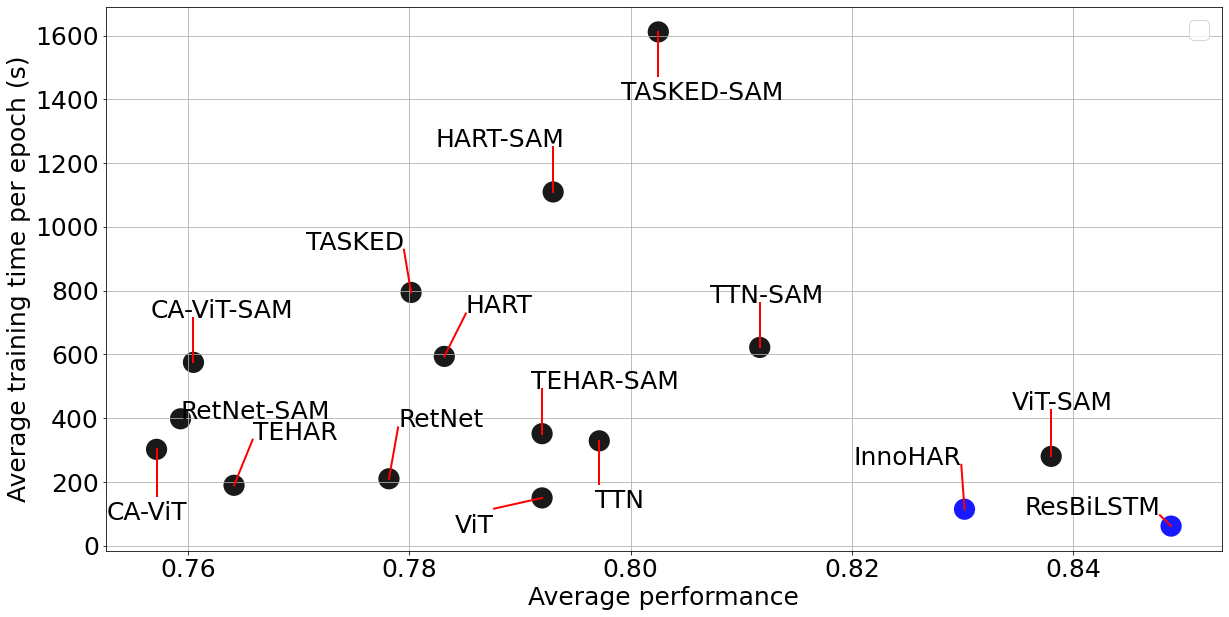}
    \caption{Average performance against training time (in seconds) across all datasets, where training time represents the duration for an entire training pass over the training set.}
    \label{fig:training}
\end{figure}

Concerning quantization (refer to Table \ref{tab:results2}), transformer-based models experience extreme degradation, reaching up to 85\% (as observed in TTN on OpportunityML). In contrast, non-transformer-based models typically remain below 5\% in terms of degradation. We find that the architectures with the highest measured sharpness (Fig. \ref{fig:hessians}) also showcase the highest performance losses. Our results align with the research by \citet{transformer_quantization_challenge}, which concludes that transformer models are challenging to quantize due to their high dynamic activation range. 

\textbf{Transformers require more computational efforts.} Fig. \ref{fig:inference} and Fig. \ref{fig:training} compare the inference time and training duration, respectively, with the performance of the architectures. To attain performance comparable to non-transformer models (i.e., a margin of within 1\%), there is approximately 20\% more computational effort during inference (ViT). The scenario becomes more pronounced in the training phase, where transformer models may take around 2.5$\times$ (ViT) to 26$\times$ (TASKED-SAM) longer. On average, the application of SAM results in an approximately 85\% increase in the training duration. Therefore, transformers are less suitable for HAR since inference in this field is often performed on resource-constrained devices. Furthermore, in the context of federated learning scenarios for HAR, where training is conducted on local devices that may also be resource-constrained, transformers may not be well-suited.

\section{Conclusions}

Challenges persist in employing transformers in sensor-based HAR. Our findings underscore that, to achieve the success seen in CV and NLP, transformer architectures necessitate a considerably larger dataset for navigating the sharp minima effectively using methods like SAM and capitalizing on the absence of inductive bias. In future research, it is also crucial to address post-quantization performance degradation and enhance robustness against adversarial attacks through the development of innovative training methods, loss functions, or architectural modifications. Future investigations in transformers for HAR should also prioritize strategies for reducing both inference and training costs. 

\bibliography{reference}
\bibliographystyle{icml2024}

\end{document}